\documentclass[letterpaper, 10 pt, conference]{ieeeconf}  

\IEEEoverridecommandlockouts                              

\usepackage{graphicx, tipa}
\usepackage{amsmath,amssymb,amsfonts}
\usepackage[caption=false]{subfig}
\usepackage[ruled,linesnumbered, noend]{algorithm2e}
\usepackage{dirtytalk}
\usepackage{hyperref}
\usepackage{gensymb}
\usepackage{tabularx}
\usepackage{multirow}
\usepackage[short]{optidef}
\usepackage{booktabs}
\usepackage{siunitx}
\usepackage{titlesec}

\title{\LARGE \bf
Adaptive Planning Framework for UAV-Based Surface Inspection in Partially Unknown Indoor Environments
}

\author{Hanyu Jin, Zhefan Xu, Haoyu Shen, Xinming Han,  Kanlong Ye, and Kenji Shimada
\thanks{Hanyu Jin, Zhefan Xu, Haoyu Shen, Xinming Han, Kanlong Ye, and Kenji Shimada are with the Department of Mechanical Engineering, Carnegie Mellon University, 5000 Forbes Ave, Pittsburgh, PA, 15213, USA. {\tt\footnotesize hanyujin@andrew.cmu.edu}}%
}

\begin{document}

\maketitle
\thispagestyle{empty}
\pagestyle{empty}

\begin{abstract}
Inspecting indoor environments such as tunnels, industrial facilities, and construction sites is essential for infrastructure monitoring and maintenance. While manual inspection in these environments is often time-consuming and potentially hazardous, Unmanned Aerial Vehicles (UAVs) can improve efficiency by autonomously handling inspection tasks. Such inspection tasks usually rely on reference maps for coverage planning. However, in industrial applications, only the floor plans are typically available. The unforeseen obstacles not included in the floor plans will result in outdated reference maps and inefficient or unsafe inspection trajectories.
In this work, we propose an adaptive inspection framework that integrates global coverage planning with local reactive adaptation to improve the coverage and efficiency of UAV-based inspection in partially unknown indoor environments. Experimental results in structured indoor scenarios demonstrate the effectiveness of the proposed approach in inspection efficiency and achieving high coverage rates with adaptive obstacle handling, highlighting its potential for enhancing the efficiency of indoor facility inspection.
\end{abstract}

\section{INTRODUCTION}
Indoor inspection has been a critical problem for infrastructure maintenance. While manual inspection is often hazardous or inefficient, Unmanned Aerial Vehicles (UAVs) have become increasingly popular in such scenes due to their agility and potential to navigate complex environments. To achieve high-quality inspection and reconstruction, UAVs must plan and execute trajectories that provide complete and consistent coverage of target surfaces, even with environmental complexity and operational constraints.

Existing UAV inspection methods typically rely on reference maps that define the inspection target \cite{FC-Planner}\cite{HCPP}\cite{CPP-2}, which can be utilized to plan global coverage trajectories that aim to minimize flight time while ensuring complete surface visibility. However, these prior maps can be outdated or incomplete due to the unmodeled static or dynamic obstacles such as temporary structures, pieces of equipment, or human workers. Such discrepancies may not only introduce navigational challenges but also result in occluded or unreachable viewpoints, leading to coverage gaps and degraded reconstruction quality.

\begin{figure}[t]
  \centering
  \includegraphics[width=\linewidth]{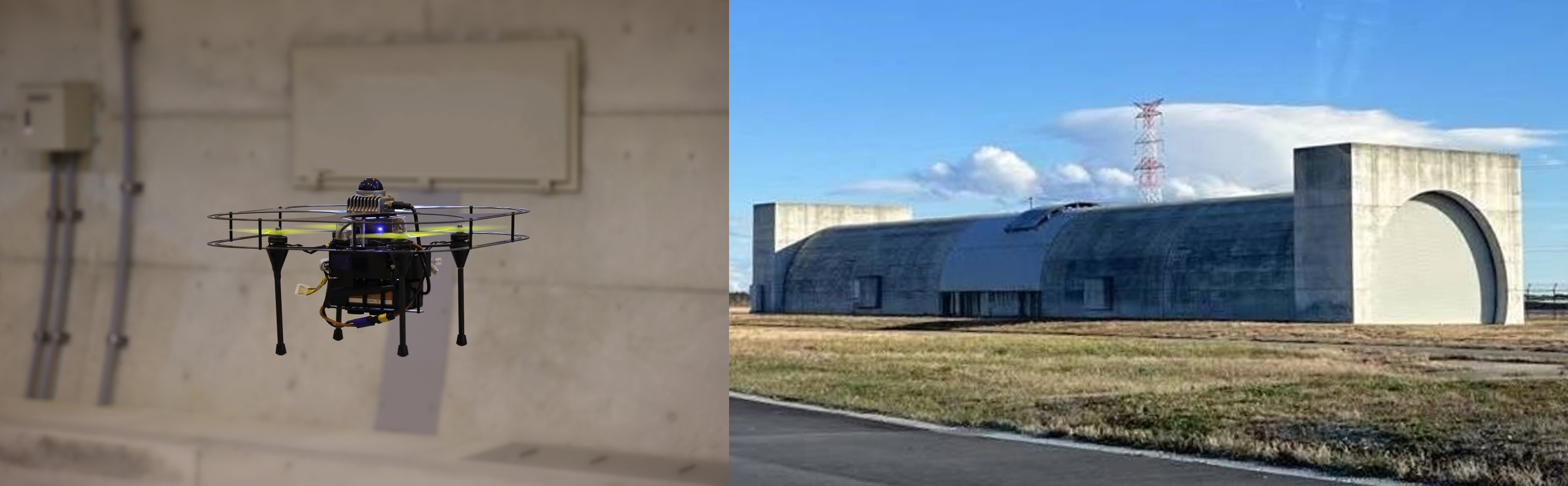}
  \caption{A UAV performing autonomous inspection in a tunnel-like indoor environment using the proposed method.}
  \label{fig:cover}
\end{figure}

To address these limitations, this work proposes a hierarchical planning framework for partially unknown indoor environments. Based on our previous work on tunnel inspection \cite{tunnel_inspection}, our system integrates a segment-based global viewpoint planner for coverage planning and a local planner that adapts the UAV trajectory and viewing angle in real time.
This integration enables the UAV to preserve inspection quality in environments with unforeseen obstacles. Figure \ref{fig:cover} shows the deployment of our framework in tunnel wall inspection. 

The main contributions of this work include:
\begin{itemize}
    \item \textbf{Surface Inspection Framework}: We introduce a  UAV inspection framework that integrates global coverage planning with local adaptability to enable efficient and complete UAV inspection in partially unknown indoor environments.
    \item \textbf{Segment-based Global Coverage Planning: }We develop a segment-based viewpoint generation and sequencing method for efficient global coverage planning.
    \item \textbf{Inspection-oriented Local Adaptive Planning: }We develop a local planner that adapts both trajectory and view angle to maximize coverage under environmental uncertainty.
    
\end{itemize}

\section{RELATED WORK}
Autonomous UAV inspection methods can be broadly categorized by their reliance on prior environmental information. Mapless approaches do not require prebuilt maps and adapt online to unknown environments. Chen et al. \cite{road_inspection} proposed a crack detection system using a Sliding Window Method (SWM) to generate trajectories from real-time images. An explore-then-exploit system in \cite{explore-then-exploit} conducts online 3D mapping before planning safe coverage paths. Similarly, in \cite{explore_and_inspect}, exploration and inspection are conducted simultaneously using heterogeneous UAV. While these methods ensure up-to-date environmental awareness, they often suffer from longer execution times and lack global optimization, leading to redundant or inefficient coverage.

In contrast, map-based approaches assume access to a reference model and solve the coverage path planning (CPP) problem through two stages: viewpoint generation and path optimization. FC-Planner \cite{FC-Planner} samples viewpoints along a geometric skeleton to reduce redundancy, while ASSCPP \cite{ASSCPP} adaptively refines sampling based on reconstruction quality. PredRecon \cite{PredRecon} leverages surface prediction for faster and more targeted viewpoint selection. BIM-informed methods \cite{BIM} incorporate robot configurations and spatial constraints to produce feasible plans in structured indoor settings. For path optimization, hierarchical methods \cite{HCPP} decompose global and local planning to improve scalability. Clustering-based frameworks \cite{clustering_cpp} segment the inspection space for efficient TSP-based routing, while view resampling \cite{vp_resample} enhances stability by smoothing redundant paths. Others integrate constraints like energy budgets using reinforcement learning \cite{RL}, enabling CPP under realistic operational limits. However, most of these approaches assume static, fully known environments, limiting their robustness in dynamic or cluttered settings.

While prior work has addressed obstacle avoidance in navigation \cite{ego_planner}\cite{vision-mpc}, few have incorporated obstacle awareness directly into the inspection objective. In contrast, our framework integrates global viewpoint planning with local obstacle-aware adaptation to improve coverage and efficiency in environments with unforeseen obstacles.

\section{METHODOLOGY}
\begin{figure}[t] 
    \centering
    \includegraphics[scale=0.46]{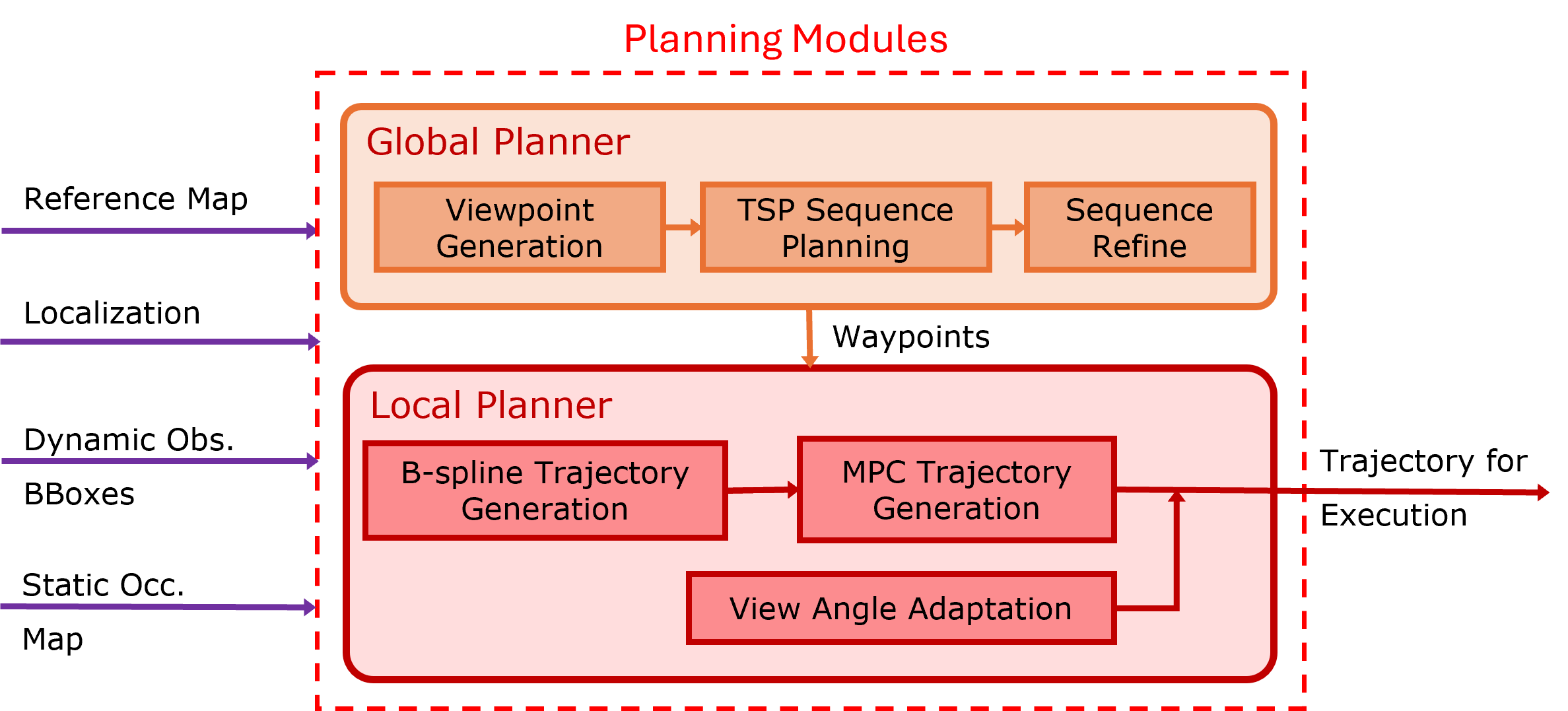}
    \caption{System overview of the proposed adaptive inspection framework. The planning module takes the reference map, localization, and perception information as input and generates a collision-free trajectory as the output. The planning module includes a global planner for viewpoint generation and sequencing and a local planner for B-spline reference trajectory generation, MPC-based tracking and collision avoidance, and adaptive view angle adjustment to ensure coverage and safe navigation during execution.}
    \label{system-overview}
\end{figure}

Figure \ref{system-overview} shows an overview of the proposed framework. Provided by a reference map $\mathcal{M}_R$, the goal is to efficiently achieve the full coverage of $\mathcal{M}_R$ using the onboard RGB-D camera for surface inspection. Localization is updated using LiDAR Inertial Odometry (LIO) 
 \cite{LIO}, and perception for both static and dynamic obstacles is provided by the onboard perception module that takes the RGB-D image and LiDAR point cloud as inputs \cite{LV-DOT}. The proposed planning method consists of two parts: a global planner for inspection viewpoint generation (Sec. \ref{sec:global_planner}) and a local planner for collision-free trajectory generation (Sec. \ref{sec:local_planner}).

\subsection{Segment-based Global Planner}\label{sec:global_planner}
The proposed global planner, as demonstrated in Alg. \ref{alg:global_planner}, takes a reference map $\mathcal{M}_R$ as input and outputs a sequence of optimized viewpoints to maximize surface inspection coverage and efficiency. 

The process begins with geometric segmentation of $\mathcal{M}_R$ using the Region Growing method (Line 2), producing a set of segments $\mathcal{S}$. This segmentation clusters the environment based on surface normals and curvature, facilitating the subsequent generation of informative viewpoints and corresponding view angles. 
For each segment $s \in \mathcal{S}$, a set of candidate viewpoints $\mathcal{V}_{\text{raw}}$ is generated to ensure comprehensive coverage (Lines 3-6). Since segmentation is conducted based on the surface normal, the viewpoints for each segment will be arranged in a linear spatial pattern. Each viewpoint $v_i$ is defined as a tuple $v_i = (p_i, \phi_i)$. $p_i$ denotes the position of the viewpoint, and $\phi_i$ is the view angle, which is initialized to be perpendicular to the principal orientation of the bounding box of the segment. 
The candidate viewpoints are then sequenced using the Lin-Kernighan-Helsgaun (LKH) solver \cite{LKH} (Line 7). Then, a post-processing with a three-stage refinement procedure is conducted to handle the outliers and optimize the consistency.
In the first stage of post-processing, a grouping is performed by mapping each viewpoint back to its source segment. Consecutive viewpoints from the same segment are merged into clusters, resulting in an initial cluster set $\mathcal{C}$ (Line 8). In the second stage, clusters with sizes below a threshold $\tau_{max}$ are considered outliers and merged with the nearest adjacent cluster from the same source segment, provided that their combined sequence remains contiguous (Lines 9-14). This yields a refined set of clusters $\mathcal{C}_{merged}$. Finally, a local reordering is performed within each cluster to minimize intra-cluster travel distance while preserving the global cluster order to generate an optimized sequence of viewpoints $\mathcal{V}_{optimized}$, which serves as the basis for downstream local planning and inspection (Lines 15-16).

\begin{algorithm}[h]
\caption{Global Planner} 
\label{alg:global_planner}
\SetAlgoNoLine
\SetKwComment{Comment}{$\triangleright$\ }{}

$\mathcal{M}_R \gets$ \text{Input Reference Map}\;

$\mathcal{S} \gets \textbf{MapSegmentation}(\mathcal{M}_R)$ \Comment{segment bounding boxes}

$\mathcal{V}_{\text{raw}} \gets \emptyset$ \ \Comment{viewpoint array}

\For{$s \in \mathcal{S}$}{
    $\mathcal{V}_s \gets \textbf{GenerateViewpoints}(s)$ \Comment{viewpoint set for each segment}
    $\mathcal{V}_{\text{raw}}.\textbf{append}(\mathcal{V}_s)$ \;
}

$\mathcal{V}_{\text{ordered}} \gets \textbf{SolveTSP}(\mathcal{V}_{\text{raw}})$ \Comment{TSP viewpoint sequence}

$\mathcal{C} \gets \textbf{Remap}(\mathcal{V}_{\text{ordered}}, \mathcal{S})$ \Comment{cluster array}

$\tau \gets \text{Initial Threshold}$;

$\tau_{max} \gets \text{Predefined Maximum Threshold}$;

\While{$\tau<=\tau_{max}$}{
    $\mathcal{C}_{merged} \gets \textbf{MergeOutlier}(\mathcal{C}, \tau)$ \Comment{merged cluster array}
    
    $\mathcal{C} \gets \mathcal{C}_{merged}$

    $\tau \gets \tau + 1$
}
$\mathcal{V}_{optimized} \gets \textbf{LocalReorder}(\mathcal{C}_{merged}, \mathcal{S})$ \Comment{output viewpoint sequence}

\textbf{return} $\mathcal{V}_{optimized}$\;

\end{algorithm}

\subsection{Inspection-oriented Local Planner}\label{sec:local_planner}
The local planner integrates a static reference trajectory planner, a dynamic obstacle avoidance planner, and a local view angle planner to maximize inspection coverage and navigation safety.

\textbf{Static Planner: }
Given a viewpoint as the navigation goal, a B-spline trajectory of order $k$ is constructed from a set of control points:
\begin{equation}
    \hat{\mathbb{P}} = \{P_1, P_2, \ldots, P_N\}, \quad P_i \in \mathbb{R}^3,
\end{equation}
where the first and last $k - 1$ control points are fixed to the start and goal positions, respectively. The trajectory generation is formulated as an unconstrained optimization problem where the optimization variables are the set $\mathbb{P}$ of the intermediate $N - 2(k - 1)$ control points, excluding the fixed start and goal. The objective is to minimize a cost function composed of weighted terms for control effort, trajectory smoothness, and static collision costs:
\begin{equation}
    J(S) = \alpha_{\text{control}} \cdot J_{\text{control}} 
    + \alpha_{\text{smooth}} \cdot J_{\text{smooth}} 
    + \alpha_{\text{static}} \cdot J_{\text{static}}.
    \label{eq:cost_function}
\end{equation}

Details of the static planner can be found in our previous work \cite{ViGO}. The static planner generates a trajectory that avoids static obstacles, which serves as a reference for the downstream dynamic planner.

\textbf{Dynamic Planner: }
Given a reference trajectory, we apply the model predictive control to generate final execution trajectories that avoid static and dynamic obstacles. The entire optimization problem can be formulated as:
\begin{mini!}[2]
    {\mathbf{x}_{0:N}, \mathbf{u}_{0:N-1}}{\sum_{k=0}^{N} {\begin{Vmatrix} \mathbf{x}_{k} - \mathbf{x}_{k}^\text{ref} \end{Vmatrix}^2 + \sum_{k=0}^{N-1} \lambda_{\mathbf{u}}\begin{Vmatrix} \mathbf{u}_{k} \end{Vmatrix}^2},}{}{} \label{mpc objective}
\addConstraint{\mathbf{x}_{0}}{=\mathbf{x}(t_{0})}{} \label{initial constraint}
\addConstraint{\mathbf{x}_{k}}{=f(\mathbf{x}_{k-1}, \mathbf{u}_{k-1})}{} \label{dynamics model}
\addConstraint{\mathbf{u}_{\text{min}} \leq}{\mathbf{u}_{k} \leq  \mathbf{u}_{\text{max}}}{} \label{control limits}
\addConstraint{\mathbf{x}_{k} \not\in \mathcal{R}_{i}}{, \forall i \in \mathbb{O}_{\text{static}} \cup \mathbb{O}_{\text{dynamic}}}{} \label{collision constraint}
\addConstraint{\forall k \in \{0, \ldots , N\}}, 
\end{mini!}
where $\mathbf{x_k} = [\mathbf{p_k},\mathbf{v_k}]^T$ and $\mathbf{u_k} = \mathbf{a_k}$ represent the robot states and control inputs with the subscript indicating the time step. The objective (Eqn. \ref{mpc objective}) is to minimize the deviation from the reference trajectory while using the least control effort.  Equation \ref{initial constraint} sets the initial state constraint based on the current robot states. The robot's dynamics model and control limits are presented by Eqns. \ref{dynamics model} and \ref{control limits}, respectively.
The collision constraints (Eqn. \ref{collision constraint}) ensure that the robot avoids collisions with static and dynamic obstacles.  Details of the dynamic planner can be found in our previous work \cite{intent_mpc}.

\textbf{View-Angle Planner: }During execution, viewpoints may become obstructed or unreachable due to unforeseen static obstacles in the environment, resulting in incomplete coverage. To mitigate this, the local planner incorporates an adaptive view angle strategy to maximize observable information in the presence of occlusions as demonstrated in Alg. \ref{alg:local_view_angle}.

The planner leverages the real-time occupancy map $\mathcal{M}$, incrementally updated from LiDAR point cloud data to identify blocked viewpoints (Line 3). Given the camera field of view (FoV) and the nominal view angle of each obstructed viewpoint $\phi_i$, the corresponding occluded regions on the reference map are inferred (Lines 4-6). Subsequently, a weight is assigned to each grid cell in the map $\mathcal{M}$ (Line 7). While all grids are initialized with the same weight, lower weights are attributed to previously scanned regions, and higher weights are assigned to occluded or unobserved areas to reflect their observation status (Lines 8-12). Then, by utilizing the current robot state and the camera field of view, the planner performs discrete 360-degree raycasting on the map $\mathcal{M}$ to evaluate the weighted visibility score across candidate view angles (Lines 13-15). The direction yielding the highest cumulative score is selected as the optimal view angle (Line 16). This adaptive mechanism enables the system to recover coverage lost to occlusions and enhances the overall robustness and efficiency of the inspection process.

\begin{algorithm}[h]
\caption{Local View Angle Adaptation}
\label{alg:local_view_angle}
\SetAlgoNoLine
\SetKwComment{Comment}{$\triangleright$\ }{}

$\mathcal{M} \gets$ \text{Input Occupancy Map}\;

$p_r \gets$ \text{Current Robot Position}\;

$\mathcal{V}_{\text{blocked}} \gets \textbf{IdentifyBlockedViewpoints}(\mathcal{M})$ \Comment{obstructed viewpoints}

\For{$v_i \in \mathcal{V}_{\text{blocked}}$}{
    $\phi_i \gets$ nominal view angle of $v_i$\;
    $\mathcal{G}_{\text{occluded}} \gets \textbf{GetOccludedRegions}(v_i, \text{FoV}, \mathcal{M}_R)$\;}
    
$\mathcal{W} \gets \textbf{InitializeGridWeights}(\mathcal{M}_R)$ \Comment{equal weight for all grids}

\For{$g \in \mathcal{M_R}$}{
    \If{$g \in \mathcal{G}_{\text{occluded}}$}{
        $\mathcal{W}[g] \gets \textbf{HigherWeight}()$ \;
    }
    \ElseIf{$g \in \mathcal{G}_{\text{scanned}}$}{
        $\mathcal{W}[g] \gets \textbf{LowererWeight}()$ \;
    }
}

$\Phi \gets$ Candidate View Angles\;

\For{$\phi \in \Phi$}{
    $S[\phi] \gets \textbf{RaycastScore}(\phi, p_r, \text{FoV}, \mathcal{W}, \mathcal{M})$ \Comment{weighted sum of visible grid scores}
}

$\phi^* \gets \arg\max_{\phi \in \Phi} S[\phi]$ \Comment{optimal view angle}

\textbf{return} $\phi^*$\;

\end{algorithm}

\section{RESULTS AND DISCUSSIONS}
\begin{figure*}[t]
  \centering
  \includegraphics[width=\linewidth]{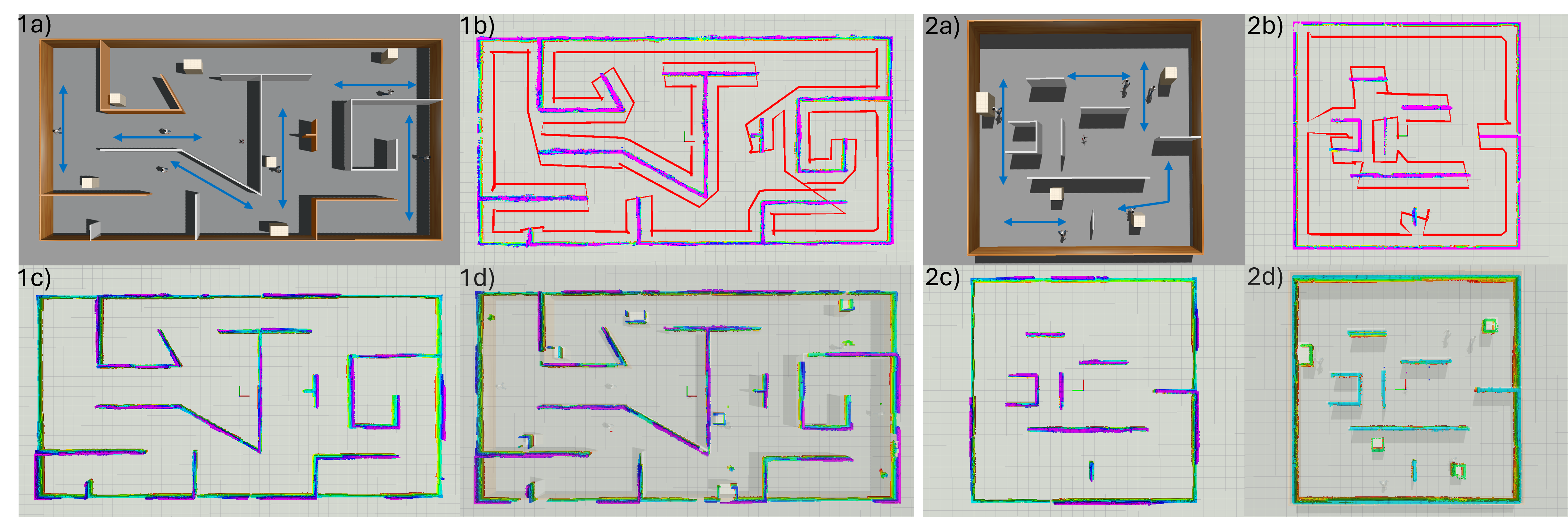}
  \caption{Simulation results in two indoor environments for evaluating inspection coverage. 1a), 2a): Top-down views of the simulated environments with static and dynamic obstacles. Blue arrows show the movement directions of the dynamic obstacles. 1b), 2b): Planned global paths on the reference maps. 1c), 2c): Voxel grids show the coverage results of camera observations in static scenes (without unforeseen static and dynamic obstacles).  1d), 2d): Voxel grids show the coverage results in the environments shown in 1a) and 2a) and are visualized over the environment.}
  \label{fig:result}
\end{figure*}
\subsection{Simulation Experiments}
To validate the proposed method, simulation experiments were conducted in two distinct indoor environments, as shown in Fig. \ref{fig:result}. Each environment includes the inspection target and the unforeseen obstacles, such as static blocks and dynamic human workers, as shown in 1a) and 2a). Provided by the floor plan, 1b) and 2b) show the reference floor plan and the planned global path. The planned paths successfully visit all target viewpoints exactly once, ensuring comprehensive coverage with minimal redundancy. Although minor collisions may occur when transitioning between distant segments, the local planner generates the trajectory in real time to maintain collision-free navigation.

Subfigures 1c) and 2c) demonstrate the inspection result in a static setting, in which the environment is the same as the provided reference map without obstacles involved. The grids indicate areas successfully captured by the onboard camera. These results show that following the planned trajectory can achieve full coverage of the inspection target. To further verify the inspection coverage under environmental uncertainty, static and dynamic obstacles (e.g., walking humans) were introduced into the scenes. As shown in 1d) and 2d), despite the discrepancy between the reference map and the real-time environment, our method successfully adapts to avoid unforeseen obstacles while preserving inspection quality, verifying its effectiveness in partially known and dynamic indoor environments.

\subsection{Physical Flight Test}
\begin{figure}
  \centering
  \includegraphics[width=\linewidth]{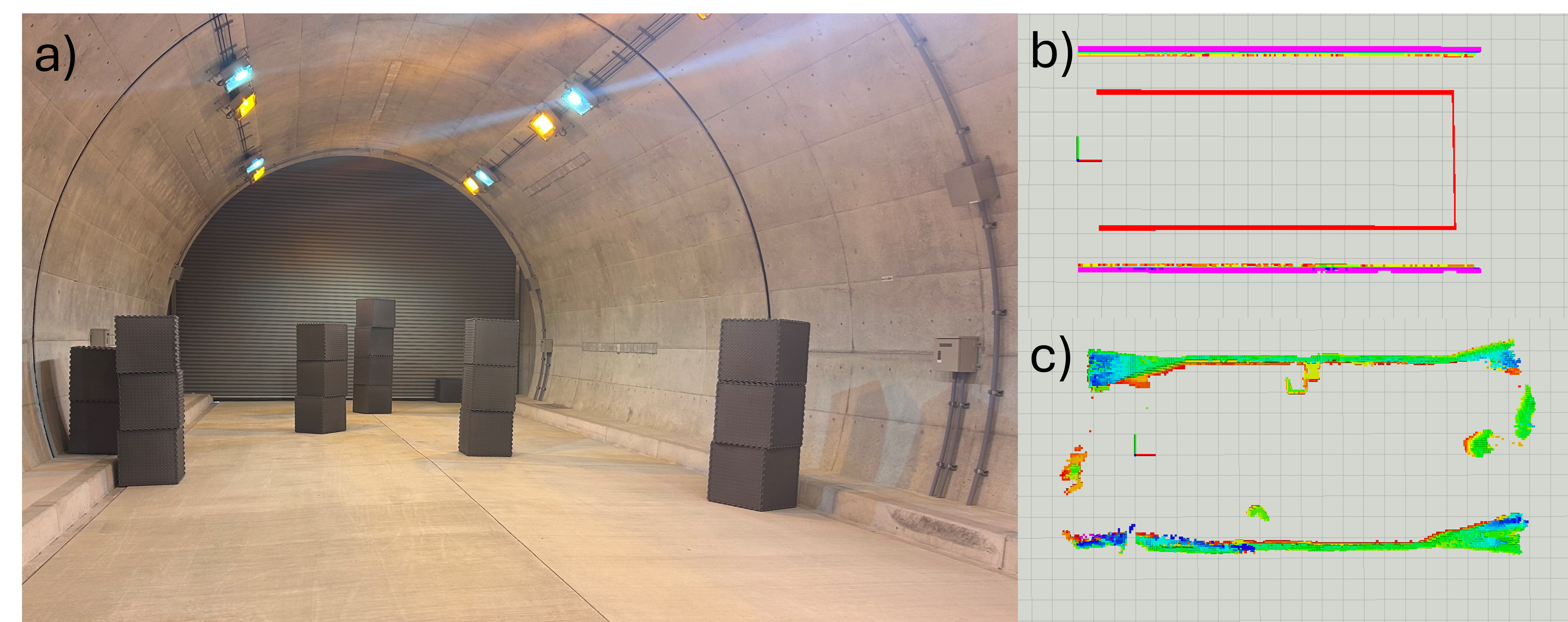}
  \caption{Real-world inspection experiment in a tunnel environment. a) The physical test environment with multiple static obstacles. b) The reference map and the planned global paths. c) Voxel grids show the coverage results in the real flight.}
  \label{fig:result_real}
\end{figure}
A physical flight test was conducted in a tunnel to validate the applicability of the proposed framework in real-world conditions. As shown in Fig. \ref{fig:result_real}, the UAV performs an autonomous inspection task in the presence of multiple static obstacles. Subfigure a) presents the actual test environment, where the tunnel walls are to be inspected and blocks were placed to simulate unknown obstacles. The reference map, as shown in subfigure b), defines the inspection target without informing the static obstacles. Despite this discrepancy, the coverage result shown in subfigure c) demonstrates that the UAV successfully maintains inspection coverage. The proposed method is able to avoid previously unknown obstacles while ensuring that the critical surfaces of interest are effectively captured, highlighting the robustness and real-world feasibility of the proposed planning framework.

\section{CONCLUSION AND FUTURE WORK}

This work presents an adaptive UAV inspection framework that integrates global viewpoint generation and coverage planning with a hierarchical local planner. The global planner generates an efficient and complete inspection path based on a reference map, while the local planner ensures safe execution by incorporating both static and dynamic obstacle avoidance as well as real-time view angle adaptation. Simulation results in structured indoor environments demonstrate that our method is able to maintain high coverage rates under environmental uncertainty. Future work will focus on processing and analyzing the collected inspection data for application in infrastructure maintenance. 

\addtolength{\textheight}{-12cm}   



\end{document}